\documentclass[10pt,twocolumn,letterpaper]{article}

\usepackage{cvpr}
\usepackage{times}
\usepackage{epsfig}
\usepackage{graphicx}
\usepackage{amsmath}
\usepackage{amssymb}
\usepackage{booktabs}
\usepackage{color, colortbl}
\usepackage{subcaption}
\usepackage{multirow}
\usepackage{pifont}
\newcommand{\cmark}{\ding{51}}%
\newcommand{\xmark}{\ding{55}}%

\newcommand{\cmmnt}[1]{}

\definecolor{Gray}{gray}{0.9}

\usepackage[pagebackref=true,breaklinks=true,letterpaper=true,colorlinks,bookmarks=false]{hyperref}

\cvprfinalcopy

\begin{document}

\title{Actor-Transformers for Group Activity Recognition}

\author{Kirill Gavrilyuk$^{1}$\thanks{This paper is the product of work during an internship at Sportlogiq.}~~~~Ryan Sanford$^{2}$~~~~Mehrsan Javan$^{2}$~~~~Cees G. M. Snoek$^{1}$\\[1mm]
\normalsize $^{1}$University of Amsterdam~~~~~$^{2}$Sportlogiq\\
{\tt\normalsize \{kgavrilyuk,cgmsnoek\}@uva.nl~~~~\{ryan.sanford, mehrsan\}@sportlogiq.com}
}

\maketitle
\begin{abstract}
This paper strives to recognize individual actions and group activities from videos. While existing solutions for this challenging problem explicitly model spatial and temporal relationships based on location of individual actors, we propose an actor-transformer model able to learn and selectively extract information relevant for group activity recognition. We feed the transformer with rich actor-specific static and dynamic representations expressed by features from a 2D pose network and 3D CNN, respectively. We empirically study different ways to combine these representations and show their complementary benefits. Experiments show what is important to transform and how it should be transformed. What is more, actor-transformers achieve state-of-the-art results on two publicly available benchmarks for group activity recognition, outperforming the previous best published results by a considerable margin.
\end{abstract}

\section{Introduction} \label{sec:intro}
The goal of this paper is to recognize the activity of an individual and the group that it belongs to~\cite{ChoiICCV2009}. Consider for example a volleyball game where an individual player \emph{jumps} and the group is performing a \emph{spike}. Besides sports, such group activity recognition has several applications including crowd monitoring, surveillance and human behavior analysis. Common tactics to recognize group activities exploit representations that model spatial graph relations between individual actors (\eg \cite{IbrahimECCV2018, QiECCV2018,WuCVPR2019}) and follow actors and their movements over time (\eg \cite{IbrahimCVPR2016, QiECCV2018, ShuCVPR2017}). The majority of previous works explicitly model these spatial and temporal relationships based on the location of the actors. We propose an implicit spatio-temporal model for recognizing group activities.

\begin{figure}[t!]
    \centering 
    \includegraphics[width=\linewidth]{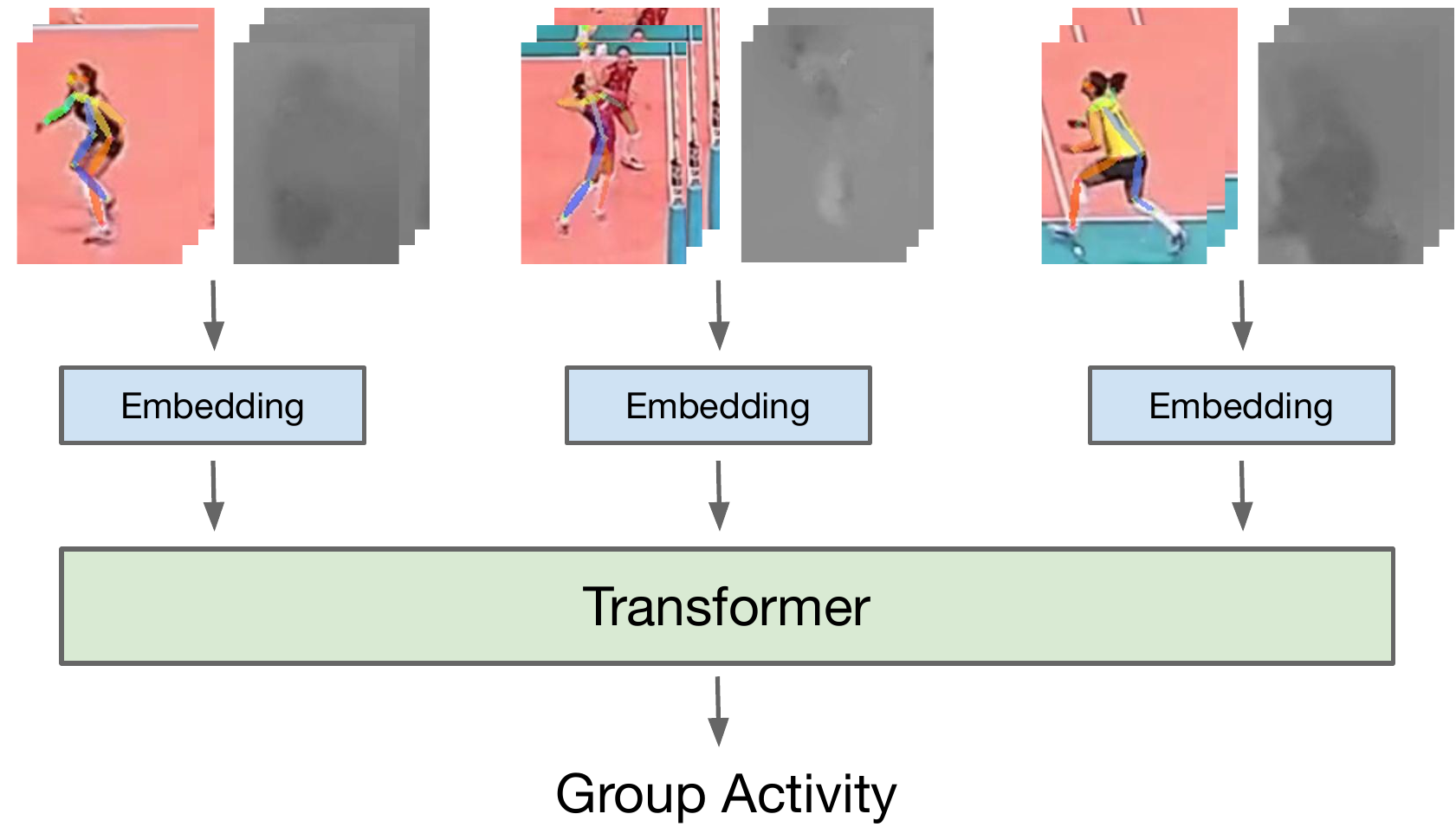} 
    \caption{We explore two complementary static and dynamic actor representations for group activity recognition. The static representation is captured by 2D pose features from a single frame while the dynamic representation is obtained from multiple RGB or optical flow frames. These representations are processed by a transformer that infers group activity.}
    \label{fig:intro}
\end{figure}

We are inspired by progress in natural language processing (NLP) tasks, which also require temporal modeling to capture the relationship between words over time. Typically, recurrent neural networks (RNN) and their variants (long short-term memory (LSTM) and gated recurrent unit (GRU)) were the first choices for NLP tasks~\cite{ChronICCV2015, MikolovICASSP2011, SutskeverICML2011}. While designed to model a sequence of words over time, they experience difficulty modeling long sequences~\cite{CollinsArxiv2016}. More recently, the transformer network~\cite{VaswaniNIPS2017} has emerged as a superior method for NLP tasks~\cite{DaiACL2019, DevlinNAACL2019, LampleArxiv2019, YangArxiv2019} since it relies on a self-attention mechanism that enables it to better model dependencies across words over time without a recurrent or recursive component. This mechanism allows the network to selectively extract the most relevant information and relationships. We hypothesize a transformer network can also better model relations between actors and combine actor-level information for group activity recognition compared to models that require explicit spatial and temporal constraints. A key enabler is the transformer's self-attention mechanism, which learns interactions between the actors and selectively extracts information that is important for activity recognition. Therefore, we do not rely on any \textit{a priori} spatial or temporal structure like graphs~\cite{QiECCV2018, WuCVPR2019} or models based on RNNs~\cite{DengCVPR2016, IbrahimCVPR2016}. We propose transformers for recognizing group activities. 

Besides introducing the transformer in group activity recognition, we also pay attention to the encoding of individual actors. First, by incorporating simple yet effective positional encoding~\cite{VaswaniNIPS2017}. Second, by explicit modeling of static and dynamic representations of the actor, which is illustrated in Figure~\ref{fig:intro}. The static representation is captured by pose features that are obtained by a 2D pose network from a single frame. The dynamic representation is achieved by a 3D CNN taking as input the stacked RGB or optical flow frames similar to~\cite{AzarCVPR2019}. This representation enables the model to capture the motion of each actor without explicit temporal modeling via RNN or graphical models.  Meanwhile, the pose network can easily discriminate between actions with subtle motion differences. Both types of features are passed into a transformer network where relations are learned between the actors enabling better recognition of the activity of the group. We refer to our approach as actor-transformers. Finally, given that static and dynamic representations capture unique, but complimentary, information, we explore the benefit of aggregating this information through different fusion strategies.

We make three contributions in this paper. First, we introduce the transformer network for group activity recognition. It refines and aggregates actor-level features, without the need for any explicit spatial and temporal modeling. Second, we feed the transformer with a rich static and dynamic actor-specific representation, expressed by features from a 2D pose network and 3D CNN. We empirically study different ways to combine these representations and show their complementary benefits. Third, our actor-transformers achieve state-of-the-art results on two publicly available benchmarks for group activity recognition, the Collective~\cite{ChoiICCV2009} and Volleyball ~\cite{IbrahimCVPR2016} datasets, outperforming the previous best published results~\cite{AzarCVPR2019, WuCVPR2019} by a considerable margin.

\section{Related Work}\label{sec:related}

\subsection{Video action recognition}
\textbf{CNNs for video action recognition.} While 2D convolutional neural networks (CNN) have experienced enormous success in image recognition, initially they could not be directly applied to video action recognition, because they do not account for time, which is vital information in videos. Karpathy~\etal~\cite{KarpathyCVPR2014} proposed 2D CNNs to process individual frames and explored different fusion methods in an effort to include temporal information. Simonyan and Zisserman~\cite{SimonyanNIPS2014} employed a two-stream CNN architecture that independently learns representations from input RGB image and optical flow stacked frames. Wang~\etal~\cite{WangECCV2016} proposed to divide the video into several segments and used a multi-stream approach to model each segment with their combination in a learnable way. Many leveraged LSTMs to model long-term dependencies across frames~\cite{DonahuePAMI2014, LiCVIU2018, NgCVPR2015,  SharmaICLR2016}. Ji~\etal~\cite{Ji2PAMI2010} were the first to extend 2D CNN to 3D, where time was the third dimension. Tran~\etal~\cite{TranICCV2015} demonstrated the effectiveness of 3D CNNs by training on a large collection of noisy labeled videos~\cite{KarpathyCVPR2014}. Carreira and Zisserman~\cite{CarreiraCVPR2017} inflated 2D convolutional filters to 3D, exploiting training on large collections of labeled images and videos. The recent works explored leveraging feature representation of the video learned by 3D CNNs and suggesting models on top of that representation~\cite{HusseinCVPR2019, WangECCV2018}. Wang and Gupta~\cite{WangECCV2018} explored spatio-temporal graphs while Hussein~\etal~\cite{HusseinCVPR2019} suggested multi-scale temporal convolutions to reason over minute-long videos. Similarly, we also rely on the representation learned by a 3D CNN~\cite{CarreiraCVPR2017} to capture the motion and temporal features of the actors. Moreover, we propose to fuse this representation with the static representation of the actor-pose to better capture exact positions of the actor's body joints.

\textbf{Attention for video action recognition.} Originally proposed for NLP tasks~\cite{BahdanauICLR2014} attention mechanisms have also been applied to image caption generation~\cite{XuICML2015}. Several studies explored attention for video action recognition by incorporating attention via LSTM models~\cite{LiCVIU2018, SharmaICLR2016}, pooling methods~\cite{GirdharNIPS2017, LongCVPR2018} or graphs~\cite{WangECCV2018}. Attention can also be guided through different modalities, such as pose~\cite{Baradel2018HumanAR, DuICCV2017} and motion~\cite{LiCVIU2018}. More recently, transformer networks~\cite{VaswaniNIPS2017} have received special recognition due to the self-attention mechanism that can better capture long-term dependencies, compared to RNNs. Integrating the transformer network for visual tasks has also emerged~\cite{GirdharCVPR2019, ParmarICML2018}. Parmar~\etal~\cite{ParmarICML2018} generalized the transformer to an image generation task, while Girdhar~\etal~\cite{GirdharCVPR2019} created a video action transformer network on top of a 3D CNN representation~\cite{CarreiraCVPR2017} for action localization and action classification. Similarly, we explore the transformer network as an approach to refine and aggregate actor-level information to recognize the activity of the whole group. However, we use representations of all actors to create query, key and values to refine each individual actor representation and to infer group activity, while ~\cite{GirdharCVPR2019} used only one person box proposal for query and clip around the person for key and values to predict the person's action.

\textbf{Pose for video action recognition.}
Most of the human actions are highly related to the position and motion of body joints. This has been extensively explored in the literature, including hand-crafted pose features~\cite{JhuangICCV2013, NieCVPR2015, WangCVPR2013}, skeleton data~\cite{DuCVPR2015, HouCSVT2018, LiuECCV2016, ShahroudyCVPR2016, SongAAAI2017}, body joint representation~\cite{CaoIJCAI2016, ChronICCV2015} and attention guided by pose~\cite{Baradel2018HumanAR, DuICCV2017}. However, these approaches were only trained to recognize an action for one individual actor, which does not generalize well to inferring group activity. In our work we explore the fusion of the pose features with dynamic representations, following the multi-stream approach~\cite{ChoutasCVPR2018, TuPR2018, ZolfaghariICCV2017} for action recognition, but we leverage it to infer group activity.

\begin{figure*}[t!]
    \centering
    \includegraphics[width=0.85\linewidth]{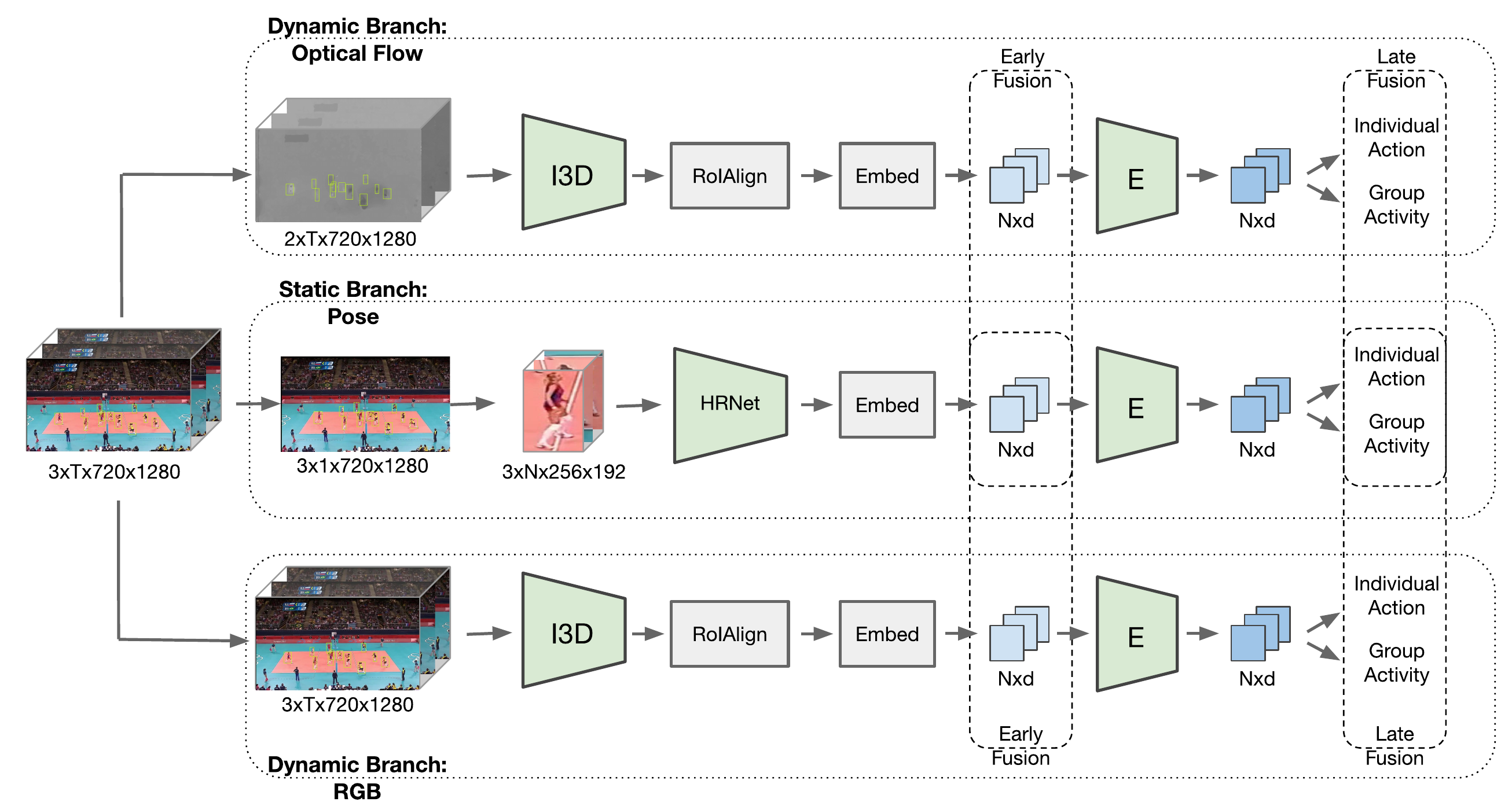}
    \caption{\textbf{Overview of the proposed model.} An input video with $T$ frames and $N$ actor bounding boxes is processed by two branches: static and dynamic. The static branch outputs an HRNet~\cite{SunCVPR2019} pose representation for each actor bounding box. The dynamic branch relies on I3D~\cite{CarreiraCVPR2017}, which receives as input either stacked RGB or optical flow frames. To extract actor-level features after I3D we apply a RoIAlign~\cite{HeICCV2017} layer. 
    A transformer encoder ($E$) refines and aggregates actor-level features followed by individual action and group activity classifiers. Two fusion strategies are supported. For early fusion we combine actor-level features of the two branches before $E$, in the late fusion we combine the classifier prediction scores. }
    \label{fig:model}
\end{figure*}

\subsection{Group activity recognition}
Group activity recognition has recently received more attention largely due to the introduction of the public Collective dataset~\cite{ChoiICCV2009} and Volleyball dataset~\cite{IbrahimCVPR2016}. Initially, methods relied on hand-crafted features extracted for each actor, which were then processed by probabilistic graphical models~\cite{AmerECCV2014, ChoiECCV2012, Choi2014PAMI, ChoiCVPR2011, HajimirsadeghiCVPR2015, LANCVPR2012, LanPAMI2012}. With the emergence of deep learning, the performance of group activity recognition has steadily increased. Some of the more successful approaches utilized RNN-type networks. Ibrahim~\etal~\cite{IbrahimCVPR2016} used LSTM to model the action dynamics of individual actors and aggregate the information to predict group activity. Deng~\etal~\cite{DengCVPR2016} integrated graphical models with RNN. Shu~\etal~\cite{ShuCVPR2017} used a two-level hierarchy of LSTMs that simultaneously minimized the energy of the predictions while maximizing the confidence. Bagautdinov~\etal~\cite{BagautdinovCVPR2017} jointly detected every actor in a video, predicted their actions and the group activity by maintaining temporal consistency of box proposals with the help of RNN. Wang~\etal~\cite{WangCVPR2017} utilizes single person dynamics, intra-group and inter-group interactions with LSTM-based model.  Li and Chuah~\cite{LiICCV2017} took an alternative approach, where captions were generated for every video frame and then were used to infer group activity. Ibrahim and Mori~\cite{IbrahimECCV2018} created a
relational representation of each person which is then used for multi-person activity recognition. Qi~\etal~\cite{QiECCV2018} proposed an attentive semantic RNN that utilized spatio-temporal attention and semantic graphs to capture inter-group relationships. Lately, studies have been moving away from RNNs. Azar~\etal~\cite{AzarCVPR2019} used intermediate representations called activity maps, generated by a CNN, to iteratively refine group activity predictions. Wu~\etal~\cite{WuCVPR2019} built an actor relation graph using a 2D CNN and graph convolutional networks to capture both the appearance and position relations between actors. Like Wu~\etal~\cite{WuCVPR2019} we also rely on actor-level representations but differently, we utilize the self-attention mechanism that has the ability to selectively highlight actors and group relations, without explicitly building any graph. Moreover, we enrich actor features by using static and dynamic representations. Similarly to~\cite{AzarCVPR2019} we build our dynamic representation with a 3D CNN.  

\section{Model}\label{sec:proposed}
The goal of our method is to recognize group activity in a multi-actor scene through enhancement and aggregation of individual actor features. We hypothesize that the self-attention mechanism provided by transformer networks is a flexible enough model that can be successfully used out-of-the-box, without additional tricks or tweaks, for the inference of the activity of the whole group given the representation of each actor.

Our approach consists of three main stages presented in Figure \ref{fig:model}: actor feature extractor, group activity aggregation and fusion. In brief, the input to our model is a sequence of video frames $F_t, t=1,..,T$ with $N$ actor bounding boxes provided for each frame where $T$ is the number of frames. We obtain the static and the dynamic representation of each actor by applying a 2D pose network on a single frame and a 3D CNN on all input frames. The dynamic representation can be built from RGB or optical flow frames, which are processed by a 3D CNN followed by a RoIAlign~\cite{HeICCV2017} layer. Next, actor representations are embedded into a subspace such that each actor is represented by a 1-dimensional vector. In the second stage, we apply a transformer network on top of these representations to obtain the action-level features. These features are max pooled to capture the activity-level features. A linear classifier is used to predict individual actions and group activity using the action-level and group activity-level features, respectively. In the final stage we introduce fusion strategies before and after the transformer network to explore the benefit of fusing information across different representations. We describe each stage in more details in the following subsections. 

\subsection{Actor feature extractor}
\label{sec:proposed:person_extractor}
All human actions involve the motion of body joints, such as hands and legs. This applies not only to fine-grained actions that are performed in sports activities (\eg \textit{spike} and \textit{set} in volleyball) but also to every day actions such as \textit{walking} and \textit{talking}. This means that it is important to capture not only the position of joints but their temporal dynamics as well. For this purpose, we utilize two distinct backbone models to capture both position and motion of joints and actors themselves. 

To obtain joints positions a pose estimation model is applied. It receives as input a bounding box around the actor and predicts the location of key joints. Our approach is independent of the particular choice of the pose estimation model. We select the recently published HRNet~\cite{SunCVPR2019} as our pose network as it has a relatively simple design, while achieving state-of-the-art results on pose estimation benchmarks. We use the features from the last layer of the network, right before the final classification layer, in all our experiments. Specifically, we use the smallest network \textit{pose\_hrnet\_w32} trained on COCO key points ~\cite{LinECCV2014}, which shows good enough performance for our task as well.  

The second backbone network is responsible for modeling the temporal dynamics. Several studies have demonstrated that 3D CNNs, with enough available data for training~\cite{TranICCV2015, CarreiraCVPR2017}, can build strong spatio-temporal representations for action recognition. Accordingly, we utilize the I3D~\cite{CarreiraCVPR2017} network in our framework since the pose network alone can not capture the motion of the joints from a single frame. The I3D network processes stacked  $F_t, t=1,..,T$ frames with inflated 3d convolutions. We consider RGB and optical flow representations as they can capture different motion aspects. As 3D CNNs are computationally expensive we employ a \textit{RoIAlign}~\cite{HeICCV2017} layer to extract features for each actor given $N$ bounding boxes around actors while processing the whole input frames by the network only once.      

\subsection{Transformer}\label{sec:proposed:transformer}
 Transformer networks were originally introduced for machine translation in~\cite{VaswaniNIPS2017}. The transformer network consists of two parts: encoder and decoder. The encoder receives an input sequence of words (source) that is processed by a stack of identical layers consisting of a multi-head self-attention layer and a fully-connected feed-forward network. Then, a decoder generates an output sequence (target) through the representation generated by the encoder. The decoder is built in a similar way as the encoder having access to the encoded sequence. The self-attention mechanism is the vital component of the transformer network, which can also be successfully used to reason about actors' relations and interactions. In the following section, we describe the self-attention mechanism itself and how the transformer architecture can be applied to the challenging task of group activity recognition in video.  

Attention $A$ is a function that represents a weighted sum of the values $V$. The weights are computed by matching a query $Q$ with the set of keys $K$. The matching function can have different forms, most popular is the scaled dot-product~\cite{VaswaniNIPS2017}. Formally, attention with the scaled dot-product matching function can be written as: 
\begin{align}
A(Q, K, V)=  \textup{softmax}(\frac{QK^T}{\sqrt{d}})V
\end{align}
where $d$ is the dimension of both queries and keys. In the self-attention module all three representations ($Q$, $K$, $V$) are computed from the input sequence $S$ via linear projections so $A(S) = A(Q(S), K(S), V(S))$.   

Since attention is a weighted sum of all values it overcomes the problem of forgetfulness over time, which is well-studied for RNNs and LSTMs~\cite{CollinsArxiv2016}. In sequence-to-sequence modeling this mechanism gives more importance to the most relevant words in the source sequence. This is a desirable property for group activity recognition as well because we can enhance the information of each actor's features based on the other actors in the scene without any spatial constraints. Multi-head attention $A_h$ is an extension of attention with several parallel attention functions using separate linear projections $h_i$ of ($Q$, $K$, $V$):
\begin{align}
A_h(Q, K, V)=  \textup{concat}(h_1, ..., h_m)W,
\label{eq:att_eqn}
\end{align}
\begin{align}
h_i = A(QW_i^Q, KW_i^K, VW_i^V)
\end{align}
Transformer encoder layer $E$ consists of multi-head attention combined with a feed-forward neural network $L$:
\begin{align}
L(X) = Linear(Dropout(ReLU(Linear(X)))
\end{align}
\begin{align}
\hat{E}(S) = LayerNorm(S + Dropout(A_h(S)))
\end{align}
\begin{align}
E(S) = LayerNorm(\hat{E}(S) + Dropout(L(\hat{E}(S))))
\end{align}
The transformer encoder can contain several of such layers which sequentially process an input $S$. 

In our case $S$ is a set of actors' features $S=\{s_i|i=1,..,N\}$ obtained by actor feature extractors. As features $s_i$ do not follow any particular order, the self-attention mechanism is a more suitable model than RNN and CNN for refinement and aggregation of these features. An alternative approach can be incorporating a graph representation as in~\cite{WuCVPR2019} which also does not rely on the order of the $s_i$. However, the graph representation requires explicit modeling of connections between nodes through appearance and position relations. The transformer encoder mitigates this requirement relying solely on the self-attention mechanism. However, we show that the transformer encoder can benefit from implicitly employing spatial relations between actors via positional encoding of $s_i$. We do so by representing each bounding box $b_i$ of the respective actor's features $s_i$ with its center point $(x_i, y_i)$ and encoding the center point with the same function $PE$ as in~\cite{VaswaniNIPS2017}. To handle 2D space we encode $x_i$ with the first half of dimensions of $s_i$ and $y_i$ with the second half. In this work we consider only the encoder part of the transformer architecture leaving the decoder part for future work.     

\subsection{Fusion}\label{sec:proposed:fusion}
The work by Simonyan and Zisserman~\cite{SimonyanNIPS2014} demonstrated the improvements in performance that can be obtained by fusing different modalities that contain complimentary information. Following their example, we also incorporate several modalities into one framework. We explore two branches, static and dynamic. The static branch is represented by the pose network which captures the static position of body joints, while the dynamic branch is represented by I3D and is responsible for the temporal features of each actor in the scene. As RGB and optical flow can capture different aspects of motion we study dynamic branches with both representations of the input video. To fuse static and dynamic branches we explore two fusion strategies: early fusion of actors' features before the transformer network and late fusion which aggregates predictions of classifiers, similar to~\cite{SimonyanNIPS2014}.  Early fusion enables access to both static and dynamic features before inference of group activity. Late fusion separately processes static and dynamic features for group activity recognition and can concentrate on static or dynamic features, separately. 

\subsection{Training objective}
\label{sec:proposed:training}
Our model is trained in an end-to-end fashion to simultaneously predict individual actions of each actor and group activity. For both tasks we use a standard cross-entropy loss for classification and combine two losses in a weighted sum:
\begin{align}
\mathcal{L}= \lambda_g\mathcal{L}_{g}(y_g, \tilde{y}_g) + \lambda_a\mathcal{L}_{a}(y_a, \tilde{y}_a)
\end{align}
where $\mathcal{L}_{g}, \mathcal{L}_{a}$ are cross-entropy losses, ${y}_g$ and ${y}_a$ are ground truth labels, $\tilde{y}_g$ and $\tilde{y}_a$ are model predictions for group activity and individual actions, respectively. $\lambda_g$ and $\lambda_a$ are scalar weights of the two losses. We find that equal weights for individual actions and group activity perform best so we set $\lambda_g=\lambda_a=1$ in all our experiments, which we detail next.


\section{Experiments}\label{sec:experiments}

In this section, we present experiments with our proposed model. First, we introduce two publicly available group activity datasets, the Volleyball dataset~\cite{IbrahimCVPR2016} and the Collective dataset~\cite{ChoiICCV2009}, on which we evaluate our approach. Then we describe implementation details followed by ablation study of the model. Lastly, we compare our approach with the state-of-the-art and provide a deeper analysis of the results. For simplicity, we call our static branch as ``Pose", the dynamic branch with RGB frames as ``RGB" and the dynamic branch with optical flow frames as ``Flow" in the following sections. 

\subsection{Datasets}\label{sec:experiments:datasets}
\textbf{The Volleyball dataset}~\cite{IbrahimCVPR2016} consists of clips from 55 videos of volleyball games, which are split into two sets: 39 training videos and 16 testing videos. There are 4830 clips in total, 3493 training clips and 1337 clips for testing. Each clip is 41 frames in length. Available annotations includes group activity label, individual players' bounding boxes and their respective actions, which are provided only for the middle frame of the clip. Bagautdinov~\etal~\cite{BagautdinovCVPR2017} extended the dataset with ground truth bounding boxes for the rest of the frames in clips which we are also using in our experiments. The list of group activity labels contains four main activities (\textit{set}, \textit{spike}, \textit{pass}, \textit{winpoint}) which are divided into two subgroups, \textit{left} and \textit{right}, having eight group activity labels in total. Each player can perform one of nine individual actions: \textit{blocking}, \textit{digging}, \textit{falling}, \textit{jumping}, \textit{moving}, \textit{setting}, \textit{spiking}, \textit{standing} and \textit{waiting}. 

\textbf{The Collective dataset}~\cite{ChoiICCV2009} consists of 44 clips with varying lengths starting from 193 frames to around 1800 frames in each clip. Every 10th frame has the annotation of persons' bounding boxes with one of five individual actions: (\textit{crossing}, \textit{waiting}, \textit{queueing}, \textit{walking} and \textit{talking}. The group activity is determined by the action that most people perform in the clip. Following~\cite{QiECCV2018} we use 32 videos for training and 12 videos for testing. 

\subsection{Implementation details}
\label{sec:experiments:implementation}
To make a fair comparison with related works we use $T=10$ frames as the input to our model on both datasets: middle frame, 5 frames before and 4 frames after. For the Volleyball dataset we resize each frame to $720\times1280$ resolution, for the Collective to $480\times720$. During training we randomly sample one frame $F_{t_p}$ from $T$ input frames for the pose network. During testing we use the middle frame of the input sequence. Following the conventional approach we are also using ground truth person bounding boxes for fair comparison with related work. We crop person bounding boxes from the frame $F_{t_p}$ and resize them to $256\times192$, which we process with the pose network obtaining actor-level features maps. For the I3D network, we use features maps obtained from \textit{Mixed\_4f} layer after additional average pooling over the temporal dimension. Then we resize the feature maps to $90\times160$ and use the RoIAlign~\cite{HeICCV2017} layer to extract features of size $5\times5$ for each person bounding box in the middle frame of the input video. We then embed both pose and I3D features to the vector space with the same dimension $d=128$. The transformer encoder uses dropout $0.1$ and the size of the linear layer in the feed-forward network $L$ is set to $256$. 

For the training of the static branch we use a batch size of 16 samples and for the dynamic branch we use a batch size of 8 samples. We train the model for 20,000 iterations on both datasets. On the Volleyball dataset we use an SGD optimizer with momentum $0.9$. For the first 10,000 iterations we train with the learning rate $0.01$ and for the last 10,000 iterations with the learning rate $0.001$. On the Collective dataset, the ADAM~\cite{KingmaICLR15} optimizer with hyper-parameters $\beta_1=0.9$, $\beta_2=0.999$ and $\epsilon=e^{-10}$ is used. Initially, we set the learning rate to 0.0001 and decrease it by a factor of ten after 5,000 and 10,000 iterations. The code of  our  model  will  be available upon publication. 

\subsection{Ablation study}
\label{sec:experiments:ablation}

\begin{table}
\centering
\begin{tabular}{cccc}
\toprule
\multirow{2}{*}{\centering\bfseries\# Layers} & \multirow{2}{*}{\centering\bfseries\# Heads} & \multicolumn{1}{p{2.0cm}}{\centering\bfseries\ Positional \\ Encoding} & \multicolumn{1}{p{1.5cm}}{\centering\bfseries\ Group \\ Activity} \\
\bottomrule
1 & 1 & \xmark & 91.0 \\
1 & 1 & \cmark & 92.3 \\
1 & 2 & \cmark & 91.4 \\
2 & 1 & \cmark & 92.1 \\
\bottomrule
\end{tabular}
\smallskip
\caption{\textbf{Actor-Transformer} ablation on the Volleyball dataset using static actor representation. Positional encoding improves the strength of the representation. Adding additional heads and layers did not materialize due to limited number of available training samples.}
\label{table:experiments:volleyball_transformer}
\end{table}

We first perform an ablation study of our approach on the Volleyball dataset~\cite{IbrahimCVPR2016} to show the influence of all three stages of the model. We use group activity accuracy as an evaluation metric in all ablations. 

\textbf{Actor-Transformer.} We start with the exploration of parameters of the actor-transformer. We experiment with the number of layers, number of heads and positional encoding. Only the static branch represented by the pose network is considered in this experiment. The results are reported in Table~\ref{table:experiments:volleyball_transformer}. Positional encoding is a viable part giving around $1.3\%$ improvement. This is expected as group activity classes of the Volleyball dataset are divided into two subcategories according to the location of which the activity is performed: \textit{left} or \textit{right}. Therefore, explicitly adding information about actors' positions helps the transformer better reason about this part of the group activity. Typically, transformer-based language models benefit from using more layers and/or heads due to the availability of large datasets. However, the Volleyball dataset has a relatively small size and the transformer can not fully reach its potential with a larger model. Therefore we use one layer with one head in the rest of the experiments. 

\begin{table}
\centering
\begin{tabular}{lccc}
\toprule
\multirow{2}{*}{\bfseries Method} & \multicolumn{1}{p{1.25cm}}{\centering\bfseries Static} & \multicolumn{2}{p{2.0cm}}{\centering\bfseries Dynamic} \\
\cmidrule(lr){2-2} \cmidrule(lr){3-4}
 & Pose & RGB & Flow \\
\toprule
Base Model & 89.9 & 89.0 & 87.8 \\
Graph~\cite{WuCVPR2019} & 92.0 & 91.1 & 89.5\\
Activity Maps~\cite{AzarCVPR2019} & - & 92.0 & 91.5 \\
\cmidrule{1-4}
Actor-Transformer (ours) & 92.3 & 91.4 & 91.5\\
\bottomrule
\end{tabular}
\smallskip
\caption{\textbf{Actor Aggregation} ablation of person-level features for group activity recognition on the Volleyball dataset. Our actor-transformer outperforms a graph while matching the results of activity maps.}
\label{table:experiments:volleyball_comparison_alternatives}
\end{table}

\textbf{Actor Aggregation.}
Next, we compare the actor-transformer with two recent approaches that combine information across actors to infer group activity. We use a static single frame (pose) and dynamic multiple frames (I3D) models as a baseline. It follows our single branch model without using the actor-transformer part, by directly applying action and activity classifiers on actor-level features from the pose and the I3D networks. The first related method uses relational graph representation to aggregate information across actors~\cite{WuCVPR2019}. We use the authors' publicly available code for the implementation of the graph model. We also use an embedded dot-product function for the appearance relation and distance masking for the position relation, which performed best in ~\cite{WuCVPR2019}. For fair comparison, we replace the actor-transformer with a graph and keep the other parts of our single branch models untouched. The second related method is based on multiple refinement stages using spatial activity maps~\cite{AzarCVPR2019}. As we are using the same backbone I3D network, we directly compare with the results obtained in~\cite{AzarCVPR2019}. The comparisons are reported in Table~\ref{table:experiments:volleyball_comparison_alternatives}. Our actor-transformer outperforms the graph for all backbone networks with good improvement for optical flow features without explicitly building any relationship representation. We match the results of activity maps~\cite{AzarCVPR2019} on optical flow and having slightly worse results on RGB. However, we achieve these results without the need to convert bounding box annotations into segmentation masks and without multiple stages of refinement.

\begin{table}
\centering
\begin{tabular}{lcc}
\toprule
\bfseries Method & \bfseries Pose + RGB & \bfseries  Pose + Flow  \\
\bottomrule
Early - summation & 91.2 & 88.5 \\
Early - concatenation & 91.8 & 89.7 \\
Late & 93.5 & 94.4 \\
\bottomrule
\end{tabular}
\smallskip
\caption{\textbf{Fusion} ablation of static and dynamic representations on the Volleyball dataset. The late fusion outperforms the early fusion approaches. }
\label{table:experiments:volleyball_comparison_fusion}
\end{table}

\textbf{Fusion.}
In the last ablation, we compare different fusion strategies to combine the static and dynamic representations of our model. For the late fusion, we set the weight for the static representation to be twice as large as the weight for the dynamic representation. The results are presented in Table~\ref{table:experiments:volleyball_comparison_fusion}. The early fusion is not beneficial for our model, performing similar or even worse than single branch models. Early fusion strategies require the actor-transformer to reason about both static and dynamic features. Due to the small size of the Volleyball dataset, our model can not fully exploit this type of fusion. Concentrating on each of two representations separately helps the model to better use the potential of static and dynamic features. Despite Flow only slightly outperforming RGB ($91.5\%$ vs. $91.4\%$), fusion with static representation has a bigger impact ($93.9\%$ vs. $93.1\%$) showing that Flow captures more complementary information to Pose than RGB. 

\begin{table}
\centering
 \resizebox{0.99\columnwidth}{!}{
\begin{tabular}{lccc}
\toprule
\multirow{2}{*}{\bfseries Method} & \multirow{2}{*}{\bfseries Backbone} & \multicolumn{1}{p{1.2cm}}{\centering\bfseries Group \\ Activity} & \multicolumn{1}{p{1.4cm}}{\centering\bfseries Individual \\ Action} \\
\bottomrule
Ibrahim~\etal~\cite{IbrahimCVPR2016} & AlexNet & 81.9 & - \\
Shu~\etal~\cite{ShuCVPR2017} & VGG16 & 83.3 & - \\
Qi~\etal~\cite{QiECCV2018} & VGG16 & 89.3 & - \\
Ibrahim and Mori~\cite{IbrahimECCV2018} & VGG19 & 89.5 & - \\
Bagautdinov~\etal~\cite{BagautdinovCVPR2017} & Inception-v3 & 90.6 & 81.8 \\
Wu~\etal~\cite{WuCVPR2019} & Inception-v3 & 92.5 & 83.0 \\
Azar~\etal~\cite{AzarCVPR2019} & I3D & 93.0 & - \\
\cmidrule{1-4}
Ours (RGB + Flow) & I3D & 93.0 & 83.7 \\
Ours (Pose + RGB) & HRNet + I3D & 93.5 & 85.7 \\
Ours (Pose + Flow) & HRNet + I3D & \textbf{94.4} & \textbf{85.9} \\
\bottomrule
\end{tabular}}
\smallskip
\caption{\textbf{Volleyball dataset comparison} for individual action prediction and group activity recognition. Our Pose + Flow model outperforms the state-of-the-art.}
\label{table:experiments:volleyball_state_of_the_art}
\end{table}

\begin{table}
\centering
 \resizebox{0.92\columnwidth}{!}{
\begin{tabular}{lcc}
\toprule
\multirow{2}{*}{\bfseries Method} & \multirow{2}{*}{\bfseries Backbone} & \multicolumn{1}{p{1.5cm}}{\centering\bfseries Group \\ Activity}\\
\bottomrule
Lan~\etal~\cite{LanPAMI2012} & None & 79.7 \\
Choi and Salvarese~\cite{ChoiECCV2012} & None  & 80.4 \\
Deng~\etal~\cite{DengCVPR2016} & AlexNet & 81.2 \\
Ibrahim~\etal~\cite{IbrahimCVPR2016} & AlexNet & 81.5 \\
Hajimirsadeghi~\etal~\cite{HajimirsadeghiCVPR2015} & None  & 83.4 \\
Azar~\etal~\cite{AzarCVPR2019} & I3D & 85.8 \\
Li and Chuah~\cite{LiICCV2017} & Inception-v3 & 86.1 \\
Shu~\etal~\cite{ShuCVPR2017} & VGG16 & 87.2 \\
Qi~\etal~\cite{QiECCV2018} & VGG16 & 89.1  \\
Wu~\etal~\cite{WuCVPR2019} & Inception-v3 & 91.0 \\
\cmidrule{1-3}
Ours (RGB + Flow) & I3D & \textbf{92.8} \\
Ours (Pose + RGB) & HRNet + I3D & 91.0  \\
Ours (Pose + Flow) & HRNet + I3D & 91.2  \\
\bottomrule
\end{tabular}}
\smallskip
\caption{\textbf{Collective dataset comparison} for group activity recognition. Our Pose + RGB and Pose + Flow models achieve the state-of-the-art results.}
\label{table:experiments:collective_state_of_the_art}
\end{table}

\subsection{Comparison with the state-of-the-art}\label{sec:experiments:stateoftheart}

\textbf{Volleyball dataset.} Next, we compare our approach with the state-of-the-art models on the Volleyball dataset in Table~\ref{table:experiments:volleyball_state_of_the_art} using the accuracy metrics for group activity and individual action predictions. We present two variations of our model, late fusion of Pose with RGB (Pose + RGB) and Pose with optical flow (Pose + Flow). Both variations surpass all the existing methods with a considerable margin: $0.5\%$ and $1.4\%$ for group activity, $2.7\%$ and $2.9\%$ for individual action recognition. It supports our hypothesis that the transformer-based model with the static and dynamic actor representations is beneficial for the group activity task. Moreover, we also compare the late fusion of RGB with optical flow representation (RGB + Flow) and achieve the same group activity accuracy as in~\cite{AzarCVPR2019} which also uses a backbone I3D network. However, we achieve these results with a much simpler approach and without requiring any segmentation annotation. Combination of all three representations gives the same performance as Pose + Flow showing that only using one dynamic representation is essential. 

\textbf{Collective dataset.} We further evaluate our model on the Collective dataset and provide comparisons with previous methods in Table~\ref{table:experiments:collective_state_of_the_art}. We use only group activity accuracy as a metric following the same approach as the related work. Interestingly, our individual branches on the Collective dataset have much more variation in their performance than on the Volleyball dataset: Flow - $83.8\%$, Pose - $87.9\%$, RGB - $90.8\%$. However, with both fused models, Pose + RGB and Pose + Flow, we achieve the state-of-the-art results, slightly outperforming the best published results of~\cite{WuCVPR2019}. We also explore the fusion of RGB and Flow representations and find that this combination performs best on the Collective dataset reaching $92.8\%$ accuracy. We hypothesize that Pose and RGB representations capture similar information that is complementary to the optical flow representation as supported by the results of Pose + RGB model which is just slightly better than RGB representation alone. We also try to combine all three representations without receiving any additional improvement over RGB + Flow. It is worth noting that with the same backbone I3D network Azar \etal \cite{AzarCVPR2019} achieve $85.8\%$ accuracy which is $7.0\%$ lower that our results showing the benefits of the transformer-based model over their activity maps approach. 

\begin{figure}[t]
\centering
\includegraphics[width=\linewidth]{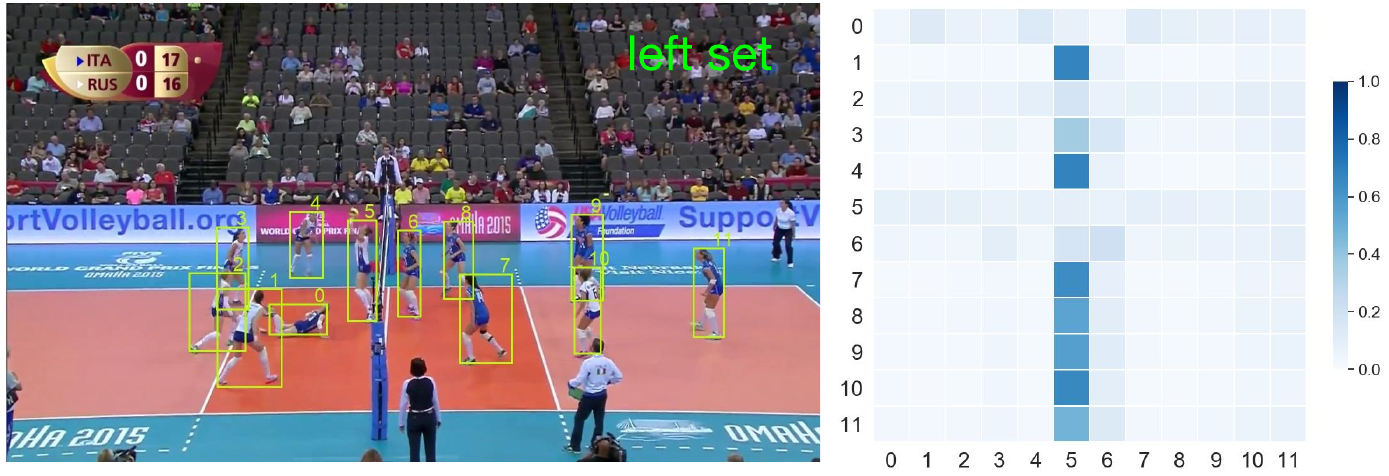}
   \caption{\textbf{Example of each actor attention} obtained by actor-transformers. Most attention is concentrated on the key actor (5) who performs \emph{setting} action which helps to correctly predict \emph{left set} group activity. Best viewed in the digital version.}
\label{fig:attn_vis_volleyball}
\end{figure}

\begin{figure}[t]
\centering
\includegraphics[width=0.7\linewidth]{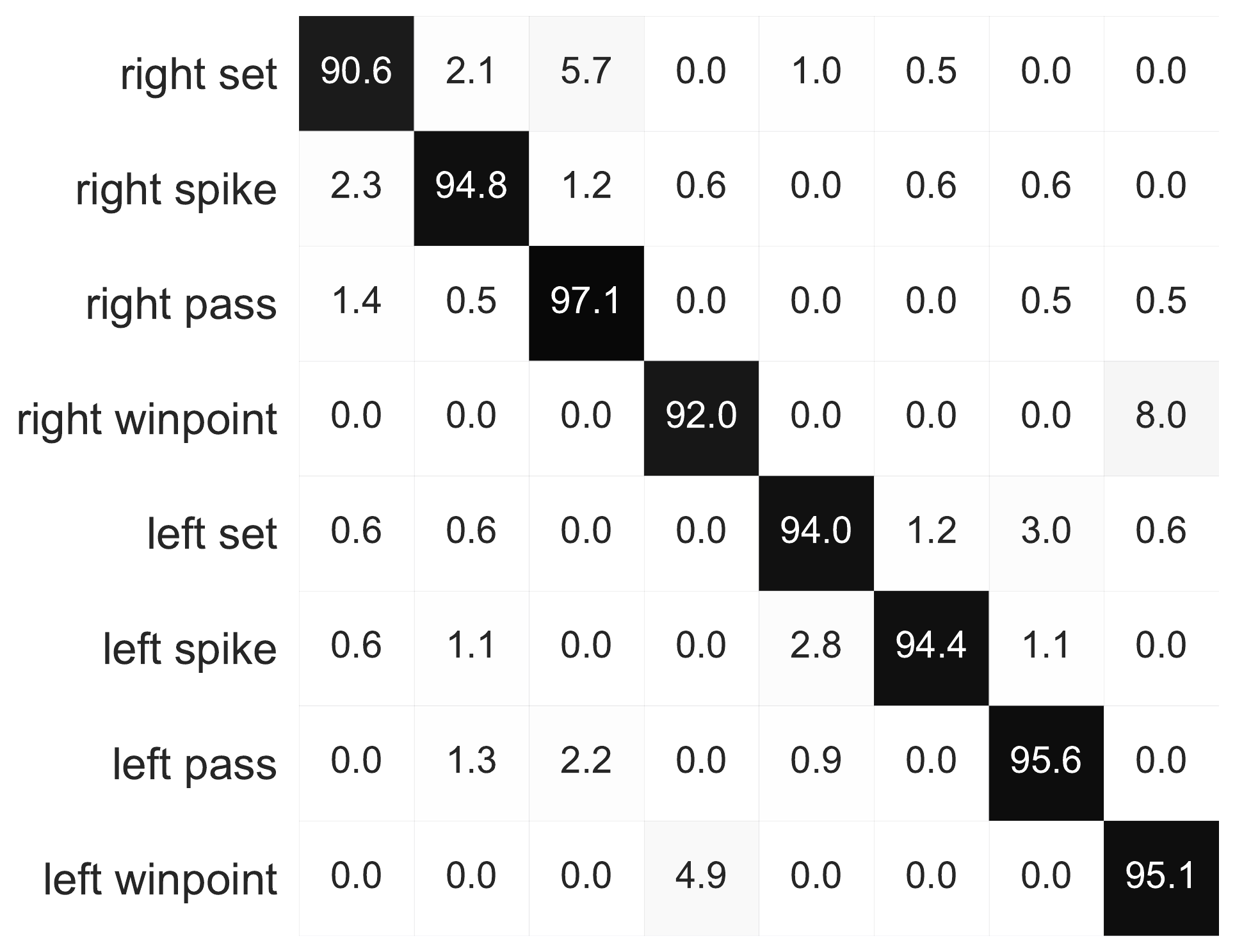}
   \caption{\textbf{Volleyball dataset confusion matrix} for group activity recognition. Our model achieves over $90\%$ accuracy for each group activity.}
\label{fig:cm_volleyball}
\end{figure}

\begin{figure}[t]
\centering
\includegraphics[width=0.7\linewidth]{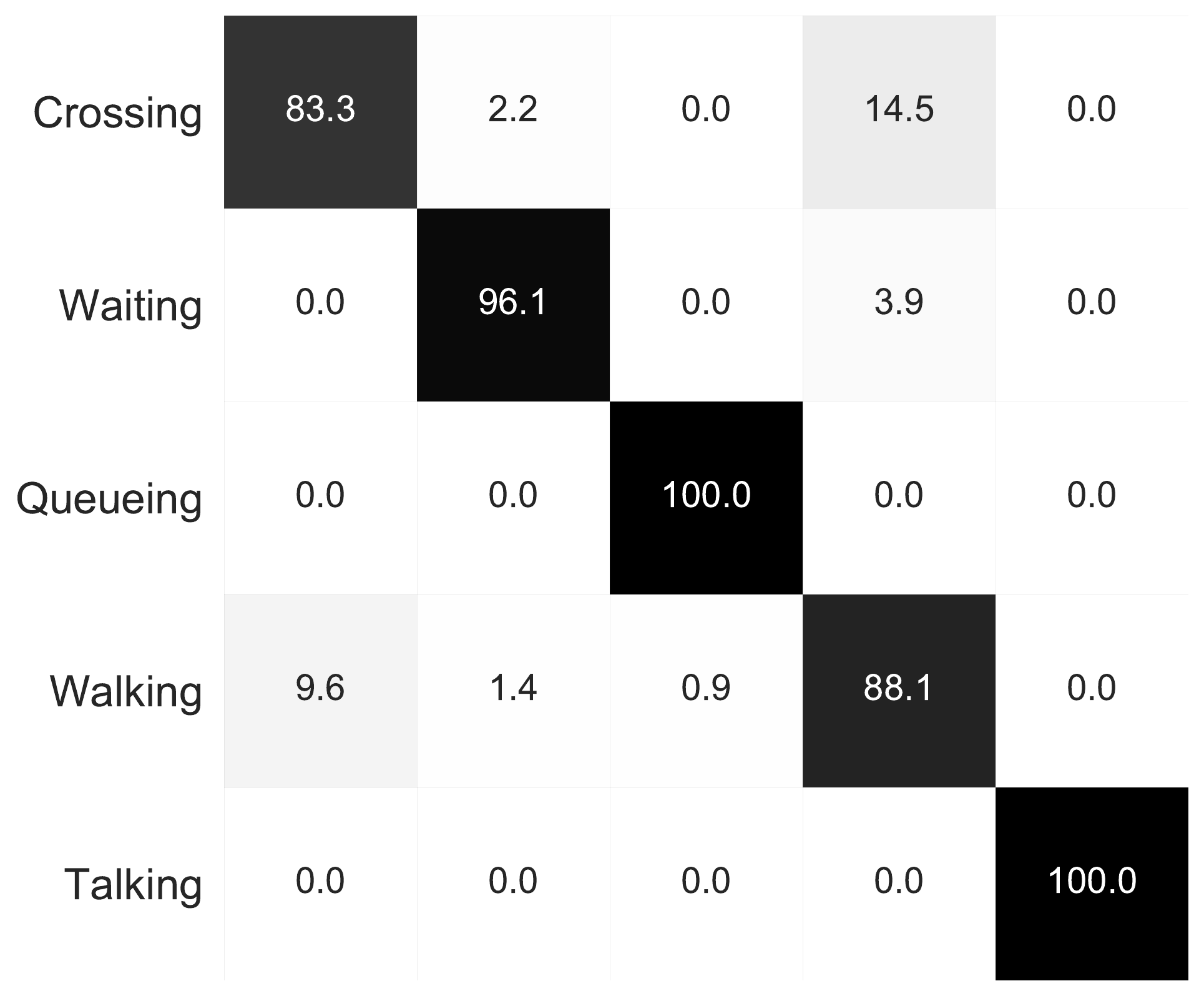}
   \caption{\textbf{Collective dataset confusion matrix} for group activity recognition. Most confusion comes form distinguishing \textit{crossing} and \textit{walking}.}
\label{fig:cm_collective}
\end{figure}

\subsection{Analysis}\label{sec:experiments:analysis}
To analyze the benefits of our actor-transformer we illustrate the attention of the transformer in Figure~\ref{fig:attn_vis_volleyball}. Each row of the matrix on the right represents the distribution of attention $A_h$ in equation ~\ref{eq:att_eqn} using the representation of the actor with the number of the row as a query. For most actors the transformer concentrates mostly on the key actor with number 5 of the \emph{left set} group activity who performs a \emph{setting} action. To further understand the performance of our model we also present confusion matrices for group activity recognition on the Volleyball dataset in Figure~\ref{fig:cm_volleyball} and the Collective dataset in Figure~\ref{fig:cm_collective}. For every group activity on the Volleyball dataset our model achieves accuracy over $90\%$ with the least accuracy for \textit{right set} class ($90.6\%$). The most confusion emerges from discriminating \textit{set}, \textit{spike} and \textit{pass} between each other despite their spatial location, \textit{left} or \textit{right}. Also, the model struggles to distinguish between \textit{right winpoint} and \textit{left winpoint}. On the Collective dataset, our approach reaches perfect recognition for \textit{queueing} and \textit{talking} classes. However, two activities, \textit{crossing} and \textit{walking}, lead to the most confusion for our model. Several works~\cite{WangCVPR2017, AzarCVPR2019} argue that \textit{crossing} and \textit{walking} are naturally the same activity as they only differ by the relation between person and street. Integrating global scene-level information potentially can help to distinguish these two activities, which we leave for future work.

\section{Conclusion}\label{sec:conclusion}
We proposed a transformer-based network as a refinement and aggregation module of actor-level features for the task of group activity recognition. We show that without any task-specific modifications the transformer matches or outperforms related approaches optimized for group activity recognition. Furthermore, we studied static and dynamic representations of the actor, including several ways to combine these representations in an actor-transformer. We achieve the state-of-the-art on two publicly available benchmarks surpassing previously published results by a considerable margin.

{\small
\bibliographystyle{ieee_fullname}
\bibliography{egbib}

\begin{thebibliography}{10}\itemsep=-1pt

\bibitem{AmerECCV2014}
Mohammed Abdel~Rahman Amer, Peng Lei, and Sinisa Todorovic.
\newblock Hirf: Hierarchical random field for collective activity recognition
  in videos.
\newblock In {\em ECCV}, 2014.

\bibitem{AzarCVPR2019}
Sina~Mokhtarzadeh Azar, Mina~Ghadimi Atigh, Ahmad Nickabadi, and Alexandre
  Alahi.
\newblock Convolutional relational machine for group activity recognition.
\newblock In {\em CVPR}, 2019.

\bibitem{BagautdinovCVPR2017}
Timur~M. Bagautdinov, Alexandre Alahi, François Fleuret, Pascal Fua, and
  Silvio Savarese.
\newblock Social scene understanding: End-to-end multi-person action
  localization and collective activity recognition.
\newblock In {\em CVPR}, 2017.

\bibitem{BahdanauICLR2014}
Dzmitry Bahdanau, Kyunghyun Cho, and Yoshua Bengio.
\newblock Neural machine translation by jointly learning to align and
  translate.
\newblock In {\em ICLR}, 2014.

\bibitem{Baradel2018HumanAR}
Fabien Baradel, Christian Wolf, and Julien Mille.
\newblock Human activity recognition with pose-driven attention to rgb.
\newblock In {\em BMVC}, 2018.

\bibitem{CaoIJCAI2016}
Congqi Cao, Yifan Zhang, Chunjie Zhang, and Hanqing Lu.
\newblock Action recognition with joints-pooled 3d deep convolutional
  descriptors.
\newblock In {\em IJCAI}, 2016.

\bibitem{CarreiraCVPR2017}
Jo{\~a}o Carreira and Andrew Zisserman.
\newblock Quo vadis, action recognition? a new model and the kinetics dataset.
\newblock In {\em CVPR}, 2017.

\bibitem{ChronICCV2015}
Guilhem Ch{\'e}ron, Ivan Laptev, and Cordelia Schmid.
\newblock P-cnn: Pose-based cnn features for action recognition.
\newblock In {\em ICCV}, 2015.

\bibitem{ChoiECCV2012}
Wongun Choi and Silvio Savarese.
\newblock A unified framework for multi-target tracking and collective activity
  recognition.
\newblock In {\em ECCV}, 2012.

\bibitem{Choi2014PAMI}
Wongun Choi and Silvio Savarese.
\newblock Understanding collective activitiesof people from videos.
\newblock {\em IEEE Transactions on Pattern Analysis and Machine Intelligence},
  36:1242--1257, 2014.

\bibitem{ChoiICCV2009}
Wongun Choi, Khuram Shahid, and Silvio Savarese.
\newblock What are they doing? : Collective activity classification using
  spatio-temporal relationship among people.
\newblock In {\em ICCV Workshops}, 2009.

\bibitem{ChoiCVPR2011}
Wongun Choi, Khuram Shahid, and Silvio Savarese.
\newblock Learning context for collective activity recognition.
\newblock In {\em CVPR}, 2011.

\bibitem{ChoutasCVPR2018}
Vasileios Choutas, Philippe Weinzaepfel, J{\'e}r{\^o}me Revaud, and Cordelia
  Schmid.
\newblock Potion: Pose motion representation for action recognition.
\newblock In {\em CVPR}, 2018.

\bibitem{CollinsArxiv2016}
Jasmine Collins, Jascha Sohl-Dickstein, and David Sussillo.
\newblock Capacity and trainability in recurrent neural networks.
\newblock {\em arXiv preprint arXiv:1611.09913}, 2016.

\bibitem{DaiACL2019}
Zihang Dai, Zhilin Yang, Yiming Yang, Jaime~G. Carbonell, Quoc~V. Le, and
  Ruslan Salakhutdinov.
\newblock Transformer-xl: Attentive language models beyond a fixed-length
  context.
\newblock In {\em ACL}, 2019.

\bibitem{DengCVPR2016}
Zhiwei Deng, Arash Vahdat, Hexiang Hu, and Greg Mori.
\newblock Structure inference machines: Recurrent neural networks for analyzing
  relations in group activity recognition.
\newblock In {\em CVPR}, 2016.

\bibitem{DevlinNAACL2019}
Jacob Devlin, Ming-Wei Chang, Kenton Lee, and Kristina Toutanova.
\newblock Bert: Pre-training of deep bidirectional transformers for language
  understanding.
\newblock In {\em NAACL-HLT}, 2019.

\bibitem{DonahuePAMI2014}
Jeff Donahue, Lisa~Anne Hendricks, Marcus Rohrbach, Subhashini Venugopalan,
  Sergio Guadarrama, Kate Saenko, and Trevor Darrell.
\newblock Long-term recurrent convolutional networks for visual recognition and
  description.
\newblock {\em IEEE Transactions on Pattern Analysis and Machine Intelligence},
  39:677--691, 2014.

\bibitem{DuICCV2017}
Wenbin Du, Yali Wang, and Yu Qiao.
\newblock Rpan: An end-to-end recurrent pose-attention network for action
  recognition in videos.
\newblock In {\em ICCV}, 2017.

\bibitem{DuCVPR2015}
Yong Du, Wei Wang, and Liang Wang.
\newblock Hierarchical recurrent neural network for skeleton based action
  recognition.
\newblock In {\em CVPR}, 2015.

\bibitem{GirdharCVPR2019}
Rohit Girdhar, Jo{\~a}o Carreira, Carl Doersch, and Andrew Zisserman.
\newblock Video action transformer network.
\newblock In {\em CVPR}, 2019.

\bibitem{GirdharNIPS2017}
Rohit Girdhar and Deva Ramanan.
\newblock Attentional pooling for action recognition.
\newblock In {\em NIPS}, 2017.

\bibitem{HajimirsadeghiCVPR2015}
Hossein Hajimirsadeghi, Wang Yan, Arash Vahdat, and Greg Mori.
\newblock Visual recognition by counting instances: A multi-instance
  cardinality potential kernel.
\newblock In {\em CVPR}, 2015.

\bibitem{HeICCV2017}
Kaiming He, Georgia Gkioxari, Piotr Doll{\'a}r, and Ross~B. Girshick.
\newblock Mask r-cnn.
\newblock In {\em ICCV}, 2017.

\bibitem{HouCSVT2018}
Yonghong Hou, Zhaoyang Li, Pichao Wang, and Wanqing Li.
\newblock Skeleton optical spectra-based action recognition using convolutional
  neural networks.
\newblock {\em IEEE Transactions on Circuits and Systems for Video Technology},
  28:807--811, 2018.

\bibitem{HusseinCVPR2019}
Noureldien Hussein, Efstratios Gavves, and Arnold~WM Smeulders.
\newblock Timeception for complex action recognition.
\newblock In {\em CVPR}, 2019.

\bibitem{IbrahimECCV2018}
Mostafa~S. Ibrahim and Greg Mori.
\newblock Hierarchical relational networks for group activity recognition and
  retrieval.
\newblock In {\em ECCV}, 2018.

\bibitem{IbrahimCVPR2016}
Mostafa~S. Ibrahim, Srikanth Muralidharan, Zhiwei Deng, Arash Vahdat, and Greg
  Mori.
\newblock A hierarchical deep temporal model for group activity recognition.
\newblock In {\em CVPR}, 2016.

\bibitem{JhuangICCV2013}
Hueihan Jhuang, Juergen Gall, Silvia Zuffi, Cordelia Schmid, and Michael~J.
  Black.
\newblock Towards understanding action recognition.
\newblock In {\em ICCV}, 2013.

\bibitem{Ji2PAMI2010}
Shuiwang Ji, Wei Xu, Ming Yang, and Kai Yu.
\newblock 3d convolutional neural networks for human action recognition.
\newblock {\em IEEE Transactions on Pattern Analysis and Machine Intelligence},
  35:221--231, 2010.

\bibitem{KarpathyCVPR2014}
Andrej Karpathy, George Toderici, Sanketh Shetty, Thomas Leung, Rahul
  Sukthankar, and Li Fei-Fei.
\newblock Large-scale video classification with convolutional neural networks.
\newblock In {\em CVPR}, 2014.

\bibitem{KingmaICLR15}
D.~P. Kingma and J. Ba.
\newblock Adam: A method for stochastic optimization.
\newblock In {\em ICLR}, 2015.

\bibitem{LampleArxiv2019}
Guillaume Lample and Alexis Conneau.
\newblock Cross-lingual language model pretraining.
\newblock {\em ArXiv}, abs/1901.07291, 2019.

\bibitem{LANCVPR2012}
Tian Lan, Leonid Sigal, and Greg Mori.
\newblock Social roles in hierarchical models for human activity recognition.
\newblock In {\em CVPR}, 2012.

\bibitem{LanPAMI2012}
Tian Lan, Yang Wang, Weilong Yang, Stephen~N. Robinovitch, and Greg Mori.
\newblock Discriminative latent models for recognizing contextual group
  activities.
\newblock {\em IEEE Transactions on Pattern Analysis and Machine Intelligence},
  34:1549--1562, 2012.

\bibitem{LiICCV2017}
Xin Li and Mooi~Choo Chuah.
\newblock Sbgar: Semantics based group activity recognition.
\newblock In {\em ICCV}, 2017.

\bibitem{LiCVIU2018}
Zhenyang Li, Kirill Gavrilyuk, Efstratios Gavves, Mihir Jain, and Cees~GM
  Snoek.
\newblock Videolstm convolves, attends and flows for action recognition.
\newblock {\em Computer Vision and Image Understanding}, 166:41--50, 2018.

\bibitem{LinECCV2014}
Tsung-Yi Lin, Michael Maire, Serge~J. Belongie, Lubomir~D. Bourdev, Ross~B.
  Girshick, James Hays, Pietro Perona, Deva Ramanan, C.~Lawrence Zitnick, and
  Piotr Doll{\'a}r.
\newblock Microsoft coco: Common objects in context.
\newblock In {\em ECCV}, 2014.

\bibitem{LiuECCV2016}
Jun Liu, Amir Shahroudy, Dong Xu, and Gang Wang.
\newblock Spatio-temporal lstm with trust gates for 3d human action
  recognition.
\newblock In {\em ECCV}, 2016.

\bibitem{LongCVPR2018}
Xiang Long, Chuang Gan, Gerard de Melo, Jiajun Wu, Xiao Liu, and Shilei Wen.
\newblock Attention clusters: Purely attention based local feature integration
  for video classification.
\newblock In {\em CVPR}, 2018.

\bibitem{MikolovICASSP2011}
Tomas Mikolov, Stefan Kombrink, Luk{\'a}s Burget, Jan {\v C}ernock{\'y}, and
  Sanjeev Khudanpur.
\newblock Extensions of recurrent neural network language model.
\newblock In {\em ICASSP}, 2011.

\bibitem{NgCVPR2015}
Joe Yue-Hei Ng, Matthew~J. Hausknecht, Sudheendra Vijayanarasimhan, Oriol
  Vinyals, Rajat Monga, and George Toderici.
\newblock Beyond short snippets: Deep networks for video classification.
\newblock In {\em CVPR}, 2015.

\bibitem{NieCVPR2015}
Xiaohan Nie, Caiming Xiong, and Song-Chun Zhu.
\newblock Joint action recognition and pose estimation from video.
\newblock In {\em CVPR}, 2015.

\bibitem{ParmarICML2018}
Niki Parmar, Ashish Vaswani, Jakob Uszkoreit, Łukasz Kaiser, Noam Shazeer,
  Alexander Ku, and Dustin Tran.
\newblock Image transformer.
\newblock In {\em ICML}, 2018.

\bibitem{QiECCV2018}
Mengshi Qi, Jie Qin, Annan Li, Yunhong Wang, Jiebo Luo, and Luc~Van Gool.
\newblock stagnet: An attentive semantic rnn for group activity recognition.
\newblock In {\em ECCV}, 2018.

\bibitem{ShahroudyCVPR2016}
Amir Shahroudy, Jun Liu, Tian-Tsong Ng, and Gang Wang.
\newblock Ntu rgb+d: A large scale dataset for 3d human activity analysis.
\newblock In {\em CVPR}, 2016.

\bibitem{SharmaICLR2016}
Shikhar Sharma, Ryan Kiros, and Ruslan Salakhutdinov.
\newblock Action recognition using visual attention.
\newblock In {\em ICLR Workshops}, 2016.

\bibitem{ShuCVPR2017}
Tianmin Shu, Sinisa Todorovic, and Song-Chun Zhu.
\newblock Cern: Confidence-energy recurrent network for group activity
  recognition.
\newblock In {\em CVPR}, 2017.

\bibitem{SimonyanNIPS2014}
Karen Simonyan and Andrew Zisserman.
\newblock Two-stream convolutional networks for action recognition in videos.
\newblock In {\em NIPS}, 2014.

\bibitem{SongAAAI2017}
Sijie Song, Cuiling Lan, Junliang Xing, Wenjun Zeng, and Jiaying Liu.
\newblock An end-to-end spatio-temporal attention model for human action
  recognition from skeleton data.
\newblock In {\em AAAI}, 2017.

\bibitem{SunCVPR2019}
Ke Sun, Bin Xiao, Dong Liu, and Jingdong Wang.
\newblock Deep high-resolution representation learning for human pose
  estimation.
\newblock In {\em CVPR}, 2019.

\bibitem{SutskeverICML2011}
Ilya Sutskever, James Martens, and Geoffrey~E. Hinton.
\newblock Generating text with recurrent neural networks.
\newblock In {\em ICML}, 2011.

\bibitem{TranICCV2015}
Du Tran, Lubomir~D. Bourdev, Rob Fergus, Lorenzo Torresani, and Manohar Paluri.
\newblock Learning spatiotemporal features with 3d convolutional networks.
\newblock In {\em ICCV}, 2015.

\bibitem{TuPR2018}
Zhigang Tu, Wei Xie, Qianqing Qin, Ronald Poppe, Remco~C. Veltkamp, Baoxin Li,
  and Junsong Yuan.
\newblock Multi-stream cnn: Learning representations based on human-related
  regions for action recognition.
\newblock {\em Pattern Recognition}, 79:32--43, 2018.

\bibitem{VaswaniNIPS2017}
Ashish Vaswani, Noam Shazeer, Niki Parmar, Jakob Uszkoreit, Llion Jones,
  Aidan~N. Gomez, Lukasz Kaiser, and Illia Polosukhin.
\newblock Attention is all you need.
\newblock In {\em NIPS}, 2017.

\bibitem{WangCVPR2013}
Chunyu Wang, Yizhou Wang, and Alan~L. Yuille.
\newblock An approach to pose-based action recognition.
\newblock In {\em CVPR}, 2013.

\bibitem{WangECCV2016}
Limin Wang, Yuanjun Xiong, Zhe Wang, Yu Qiao, Dahua Lin, Xiaoou Tang, and
  Luc~Van Gool.
\newblock Temporal segment networks: Towards good practices for deep action
  recognition.
\newblock In {\em ECCV}, 2016.

\bibitem{WangCVPR2017}
Minsi Wang, Bingbing Ni, and Xiaokang Yang.
\newblock Recurrent modeling of interaction context for collective activity
  recognition.
\newblock In {\em CVPR}, 2017.

\bibitem{WangECCV2018}
Xiaolong Wang and Abhinav Gupta.
\newblock Videos as space-time region graphs.
\newblock In {\em ECCV}, 2018.

\bibitem{WuCVPR2019}
Jianchao Wu, Limin Wang, Li Wang, Jie Guo, and Gangshan Wu.
\newblock Learning actor relation graphs for group activity recognition.
\newblock In {\em CVPR}, 2019.

\bibitem{XuICML2015}
Kelvin Xu, Jimmy Ba, Ryan Kiros, Kyunghyun Cho, Aaron~C. Courville, Ruslan
  Salakhutdinov, Richard~S. Zemel, and Yoshua Bengio.
\newblock Show, attend and tell: Neural image caption generation with visual
  attention.
\newblock In {\em ICML}, 2015.

\bibitem{YangArxiv2019}
Zhilin Yang, Zihang Dai, Yiming Yang, Jaime~G. Carbonell, Ruslan Salakhutdinov,
  and Quoc~V. Le.
\newblock Xlnet: Generalized autoregressive pretraining for language
  understanding.
\newblock {\em ArXiv}, abs/1906.08237, 2019.

\bibitem{ZolfaghariICCV2017}
Mohammadreza Zolfaghari, Gabriel~L. Oliveira, Nima Sedaghat, and Thomas Brox.
\newblock Chained multi-stream networks exploiting pose, motion, and appearance
  for action classification and detection.
\newblock In {\em ICCV}, 2017.

\end{thebibliography}
}

\end{document}